# Toward Educator-focused Automated Scoring Systems for Reading and Writing

Mike Hardy
University of California, Berkeley
Hardy.mike@berkeley.edu

## Abstract

This paper presents methods for improving automated essay scoring with techniques that address the computational trade-offs of self-attention and document length. To make Automated Essay Scoring (AES) more useful to practitioners, researchers must overcome the challenges of data and label availability, authentic and extended writing, domain scoring, prompt and source variety, and transfer learning. This paper addresses these challenges using neural network models by employing techniques that preserve essay length as an important feature without increasing model training costs. It introduces techniques for minimizing classification loss on ordinal labels using multi-objective learning, capturing semantic information across the entire essay using sentence embeddings to use transformer architecture across arbitrarily long documents, the use of such models for transfer learning, automated hyperparameter generation based on prompt-corpus metadata, and, most importantly, the use of semantic information to provide meaningful insights into student reading through analysis of passage-dependent writing resulting in state-of-the-art results for various essay tasks.

## 1 Introduction

Assessments measuring student learning need to increase in validity—accurately measuring intended learning. While standardized, multiple-choice tests have increased statistical reliability, writing remains a more authentic measure of student learning. However, hand-scoring essays is both expensive and time-consuming, leading to increased emphasis on and subsequent leaps in Automated Essay Scoring (AES) and Short Answer Scoring (SAS) in the last two decades.

However, the market and research for AES has been driven primarily by companies who do extensive essay grading and primarily focuses on producing a reliable holistic—or overall—scores (see Ke et al., 2019). While these scores serve the purposes of standardized testing, holistic scores are less useful to educators working to remediate immediate student needs, and, for educators in public schools, AES services are typically unavailable entirely.

Simultaneously, there is a great desire nationally among education agencies, to discard the traditional summative assessment model as enshrined by No Child Left Behind (NCLB) for a more authentic and instructionally relevant testing experience. The most desirable specifications for assessments of English Language Arts (ELA) would be simultaneously writing-based and standardized with immediate feedback available to students and educators, passage-dependent for assessing reading, 'warm' or 'hot'[1] passages-prompt combinations aligned to high-quality instructional materials to minimize schema advantages and maximize validity, cover the range of the state standards and skills as required in the Elementary and Secondary Education Act of 1965 (ESEA), account for student development and growth throughout the course of the school year, and reduce or eliminate the educational 'autopsy' of summative testing. Many assessment practitioners believe that it is impossible to meet all those requirements, but powerful models like GPT-3 (Brown et al., 2020) are forcing the industry to rethink limitations to how natural language can be evaluated.

The work of this paper is to refocus research in AES to be educator-focused and pedagogically aligned by introducing methodologies that value the long-term needs of educators and students. Teachers want a free, useful, and insightful automatic scorer for student writing, and students want meaningful instruction and feedback in order to improve. AES' becoming more useful and insightful requires including domain-specific scores, in addition to holistic scores, that represent pedagogically aligned writing traits that match instructional needs of teachers and the learning needs of students.

This is a massive challenge, as this would require zero-shot learning on novel prompts of arbitrary length, which may be source-dependent and would need to be scored across various writing traits accounting for the writing development progression found by student age within K-12 schools. Meeting this challenge requires a re-focus of current AES research trends to work toward that ultimate outcome.

---

[1] 'Hot' assessment items are where educators and/or students are given all or part of an item before the assessment. 'Warm' assessment items are where the topic, theme, or focus of the item is shared beforehand.

To better serve educators and learners, some of the most immediate challenges for AES include data availability, authentic and extended writing, domain scoring, transfer learning, and prompt and source variety.

## 2 Related Work

This section will detail approaches to solving some of the challenges in public educator-focused essay scoring.

### 2.1 Corpora and Labeling

Due to the need to protect the educational information of minors, there is limited data available to the public. This provides a great advantage to companies to provide grading services, as they jointly share a monopoly on essay grading data. For essay grading, only one dataset representing K-12 students is publicly available: the Hewitt Foundation's ASAP-AES dataset[2] used in the 2012 Kaggle competition of the same name (Table 1). A similar dataset for shorter answers, ASAP-SAS[3] was also released through the Hewitt Foundation, which includes five passage-dependent shorter responses for English Language Arts. The ASAP-AES dataset is annotated holistically, with only two of the eight essays containing trait annotations available beyond holistic scoring, which were not all evaluated for the competition and offer different traits using different scoring and scaling criteria. While using the same data source is helpful for benchmark comparison, it significantly restricts the robustness of model transferability and increases the bias found within the sample when developing AES methodologies.

Additionally, the score ranges and grading criteria are different for different prompts, making it challenging to effectively address gaps across prompts. Due to computational restrictions or to avoid the challenges of unusual classifications, many researchers create models on only part of the ASAP set (Zhang et al., 2018; Nadeem, et al., 2019, Mayfield et al., 2020)

Nearly all existing AES systems were developed for holistic scoring, with trait-specific scoring beginning only in 2004. Primarily using researcher-generated estimations, features, or ratings, a few studies on single traits have shown the difficulty of the task, with traits such as organization (Persing et al., 2010; Taghipour, 2017), prompt adherence (Persing and Ng, 2014), thesis clarity (Persing and Ng, 2013; Ke et al., 2019), argument strength (Persing and Ng, 2015; Taghipour, 2017; Carlile et al., 2018), coherence (Somasundaran et al., 2014), and stance (Persing and Ng, 2016), and narrative quality (Somasundaran et al., 2018). Responding to the challenge of obtaining trait-labeled datasets, Mathias et al. (2016), produced trait labels for the remaining six essays to create the ASAP++ dataset. These trait labels are not gold-standard (as they received ratings from only one reviewer), but their effort is a worthy contribution, as it is followed up with a study (Mathias et al., 2020) replicating certain SOTA models across multiple traits. Through its imperfections, this label augmentation heightens the need for scoring models that support transfer learning, so when a gold-standard, trait-labeled dataset becomes available, we have algorithms that can fully take advantage of that learning.

Another noteworthy limitation of the ASAP dataset is that the essays have already been standardized to a degree, as they do not contain any paragraph information and have gone through significant preprocessing to remove named entities and most capitalized words. While this was a precaution to protect minors from potentially identifiable information, it limits the long-term ability of this dataset to produce more authentic models.

| Essay | Grade | Len | Score Range | Essay Count | Source Dep | Essay Description |
|---|---|---|---|---|---|---|
| **1** | 8 | 350 | 2-12 | 1785 | N | Persuasive Letter about Technology Use |
| **2** | 10 | 350 | 1-6 | 1800 | N | Persuasive Essay about Library Censorship |
| **3** | 10 | 150 | 0-3 | 1726 | Y | Source-dependent Analysis of Setting |
| **4** | 10 | 150 | 0-3 | 1772 | Y | Source-dependent Analysis of Author's Purpose |
| **5** | 8 | 150 | 0-4 | 1805 | Y | Source-dependent Analysis of Mood |
| **6** | 10 | 150 | 0-4 | 1800 | Y | Source-dependent Demonstration of comprehension of Text |
| **7** | 7 | 250 | 0-30 | 1730 | N | Narrative about Patience |
| **8** | 10 | 650 | 0-60 | 918 | N | Narrative about Laughter |

Table 1: Overview of the ASAP-AES dataset from the Hewitt Foundation including prompt, grade-level of authors, average length of essay in words, range of possible scores, number of essays per prompt, and essay description.

---

[2] https://www.kaggle.com/c/asap-aes/

[3] https://www.kaggle.com/c/asap-sas

In short, AES desperately needs labeled data, and, until that is available, on-going research needs to prioritize methods that extract more meaningful information from limited samples.

## 2.2 AES Approaches

AES methods have typically relied on handcrafted features (Larkey, 1998; Foltz et al., 1999; Attali and Burstein, 2006; Dikli, 2006; Wang and Brown, 2008; Chen and He, 2013; Somasundaran et al., 2014; Yannakoudakis et al., 2014; Phandi et al., 2015), a technique that can be labor-intensive and, for domain scoring and source-dependent prompts, less transferrable. This has made the creation of AES systems largely limited to assessment companies and has restricted the ability to capture meaning and content, which would be required for modeling challenging writing domains.

The extensive use of feature engineering (Zesch, Wojatzki, and Scholten-Akoun 2015) has been used to produce the most commonly cited baseline for the ASAP dataset, a model called the "Enhanced AI Scoring Engine" (EASE) which was the best open-source system from the Kaggle competition, achieving third place overall (Phandi et al., 2015).

Following this trend, the most successful publicly available models utilize each full essay but simultaneously reduce the ability to recover levels of meaning that would support natural language inference or question-answering required for robust domain-scoring. Cozma et al (2018) recorded state-of-the-art results across the full ASAP dataset, but employed character-level 15-grams as string kernels and pretrained and grouped word embeddings, providing the only source of semantic information. While producing impressive results, the model does not meaningfully move practitioner-focused research forward.

Extended writing can help reveal student learning in ways that short responses cannot. However, dealing with length and extracting meaningful inference has been challenging for researchers, and the approach has been to choose either utilizing the full essay length or trying to extract higher-level semantic features.

To reduce expensive feature-engineering, neural approaches to automated essay scoring began in 2016 using corpus-specific word embeddings with LSTMs (Alikaniotis et al., 2016; Mim et al., 2019) and CNNs (Dong and Zhang, 2016; Dasgupta et al., 2018), as well as combined LSTM + CNN ensembles (Taghipour and Ng, 2016). The latter paper has become the most widely cited deep-learning AES approach, and the model evaluation methods are those still used by AES researchers, as the test dataset for the ASAP competition is not publicly available. Taghipour and Ng took one-hot word vectors as input, using a convolution layer to extract n-gram level dependencies before processing by an LSTM to capture longer distance dependencies. Using LSTMs has been very useful to capture the logical and semantic flow of essays, culminating in Tay et al. (2018) implementing an LSTM model with an attention window across its hidden states, helping to address information loss across longer essays.

Transformer-based models have added great power to natural language processing tasks (Vaswani et al., 2017) by using attention instead of RNNs, eliminating temporal information loss. These breakthroughs have resulted in powerful language representations, most notably BERT (Devlin et al., 2019). AES has seen various BERT and XLNet implementations (Rodriguez et al, 2019, Yang et al.,2019, Mayfield and Black, 2020) including the current deep-learning methods SOTA (Liu et al., 2019). The method employed by Liu et al. (2019) team is very complex and computationally expensive, utilizing three separate implementations of BERT embeddings topped with LSTMs across these values and separate BERT values for the prompt, before being combined with a second model, which employs feature engineering (seemingly losing some of the purpose of neural models). Despite the extensive power of the model, it still did not outperform the model of Cozma et al. (2018).

The failure of BERT and the transformer revolution to revolutionize essay grading is due to the computational requirements of transformers and the restrictions of BERT. As Vaswani et al. (2017) note, self-attention mechanisms increase in computational complexity proportionally with the square of the length of the sequence and the dimensions of the model. BERT limits this computational expense by restricting its tokenized input to 512 characters, resulting in every BERT implementation truncating essays arbitrarily to fit the sequence length constraint. When students are demonstrating learning through writing, not all their most important writing comes at the beginning of an essay nor can be decontextualized by restricting vocabulary. Until this paper, there has not yet been an AES model that employs self-attention without essay truncation across the entirety of the ASAP data set. AES research should explore models with more efficient self-attention of arbitrarily long documents like Beltagy, et al. (2020), which works to reduce the cost of the attention matrix computations, or begun by Nadeem et al. (2019) who implement a hierarchical attention model (Yang et al., 2016; Kowsari et al., 2017). Reducing computation or start-up costs to model training is important in supporting the objective of providing a free and open-source resource to teachers.

## 3 Contributions

This paper builds upon previous research to improve AES systems for practitioners, by employing techniques that use the full essay with self-attention without increasing computation by the square of the length. New contributions to AES using the ASAP dataset include:

- First application of ordinal classification structures (Chen and He, 2013) to deep learning models.
- First use multiple objective functions minimize classification loss essay scores
- First use of semantic sentence-level embeddings via the Universal Sentence Encoder (USE) (Cer et al., 2018) and Sentence-BERT (Riemers and Gurevych, 2019) as the atomic vectors supplied to a deep network to capture semantic information across entire essays, allowing for processing of much longer documents.
- Automatic hyperparameter generation based on essay prompt and corpus metadata. While this data is used extensively in statistical language models and handcrafted features, it has not been applied to the task of automatically tuning hyperparameters.
- First use of multi-head attention (Vaswani et al., 2017) across sentence embeddings for AES.
- First use generalized transfer modeling of essay scoring.

## 4 Models

All experiments used USE sentence embeddings as a scalable and transferrable method for capturing complex content information from essays.

- **LSTM**: a simple LSTM model (Hochreiter et al., 1997), with two dense layers prior to classification, was used with USE embedding inputs for comparison to baseline and for simple experimentation.
- **MHA**: Multi-head Attention, shown in figure 1, based on the transformer model from Vaswani et al. (2017)[4], which inputs utilizing only a classifier vector from each essay in a single dense layer feed forward classifier and Luong multiplicative attention (Luong et al., 2015).

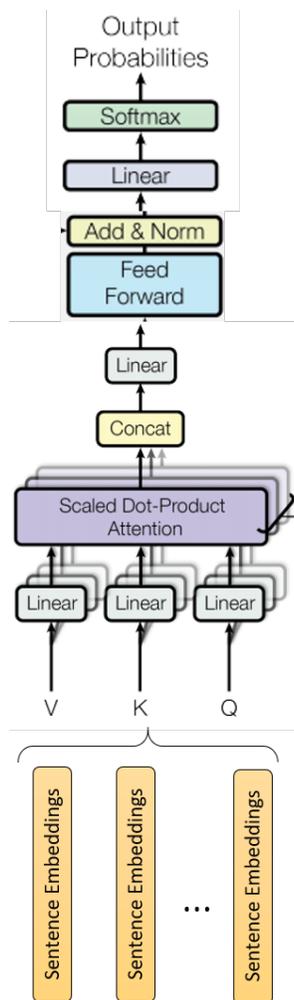

Figure 1: Simple Multi-head Attention Classifier with USE sentence embeddings.

- **2MHA**: Two-layer Multi-head Attention, using the CLS vector for the single feed forward classifier.
- **MHA+BLSTM**: This model takes multiple layers of the MHA model and then uses a bidirectional LSTM to aggregate the attended features temporally prior to passing the values to a dense classifying network. This model intends to maximize the temporal information in longer essays by capturing features across all hidden outputs of the multi-head attention layers with both combined LSTM and feed forward networks.
- **PD-S-BERT**: Passage-dependent Sentence-BERT includes the passages and quick essay statistics as inputs to allow a single algorithm to perform across all passage-dependent essays. This model was trained using curricular learning (Graves et al., 2016), with the intent for future research around few-shot and zero-shot learning. The full model for PD-S-BERT can be found in Figure 2.

## 5 Evaluation

Since the initial Kaggle competition, AES systems typically use Quadratic Weighted Kappa (QWK) on each essay set separately for evaluation; however, other measures include Pearson's correlation, Spearman's correlation, Kendall's Tau, and Cohen's Kappa. However, not all evaluation methods are practically meaningful. For example, the first LSTM model for AES (Alikaniotis *et al.*, 2016) used Cohen's Kappa across all essays and scoring systems, producing a score that does not actually "surpass similar state-of-the-art systems" in real-world results, as the paper claims. This is because essay sets that are on different scoring ranges cannot be evaluated without taking into consideration the actual possible earnable ratings. Since that time, other researchers (e.g. Dasgupta et al., 2018) have used that same metric.

Since (Taghipour and Ng, 2016), all analyses using this dataset have used both QWK and a 5-fold cross validation (60% train/20% dev/20% test) to evaluate models, since the test set used in the competition is not publicly available. As a baseline, Taghipour and Ng (2016) utilize Bayesian linear ridge regression (BLRR) as reported in (Phandi et al., 2015), as implemented in the EASE system. For comparability, this paper follows all the same evaluation practices used with this dataset, with the exception that experiments were conducted that reduced the training and validation sets to test various models' abilities with even more limited data.

---

[4] Code was modified from Vaswani et al., (2017) found at www.tensorflow.org/tutorials/text/transformer

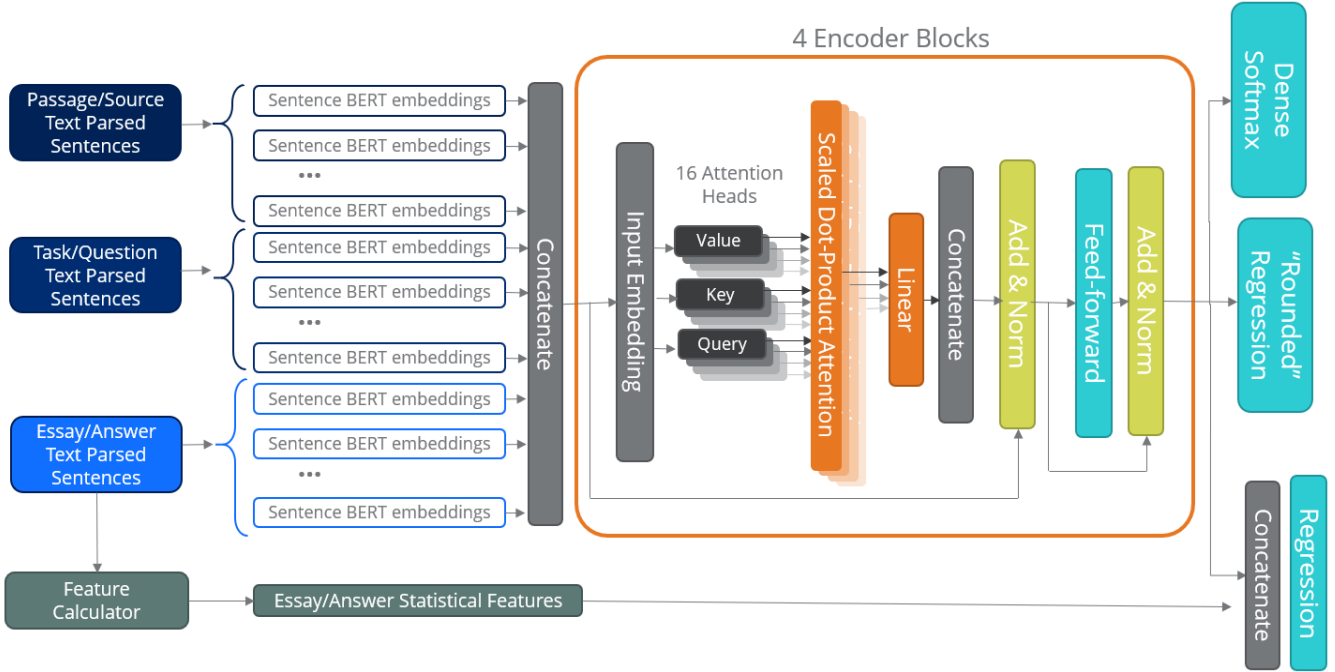

*Figure 2: Full transfer-learning model architecture for PD-S-BERT model*

## 6 Experiments

In pursuit of educator-focused AES, several experiments were conducted to facilitate future research.

### 6.1 Setting of Hyperparameters

Aside from some values explicitly discussed below and in the interest of rapid transferability of models to novel scenarios, hyperparameters were chosen through defining relationships from known essay set metadata. Each hyperparameter that was tuned in this way is listed in Table 2 with the corresponding metadata values of the essay sets.

| Hyperparameter | Metadata used to Define |
|---|---|
| Loss Weight (see 6.2) | # of classes |
| Dropout | # of observations, # of classes |
| Dimension of feed forward layer | # of classes |
| # of attention heads | # of classes, prompt type |
| Batch size | # of observations, # of classes |
| Epochs | # of observations, # of classes |
| Callback Patience | # of observations, # of classes |

Table 2: Hyperparameter tuning based on AES-specific essay metadata.

Various experiments were conducted on the appropriate learning rate, however, the final learning rate used is defined using a custom schedule taken from Vaswani et al. (2017):

$$lrate = d_{model}^{-0.5} * min(step\_num^{-0.5}, step\_num * warmup\_steps^{-1.5})$$

### 6.2 Categorical and Regression Classification

While every other deep-learning AES study after (Taghipour and Ng, 2016) treats autograding as a regression problem, correlation may not be as meaningful for essay scoring with fewer than 6 labels. The use of regression as the only loss function for deep learning AES models unnecessarily restricts the amount of information contained in labels. For essays with many possible labels (like essays 7 and 8), regression was more powerful. However, state-of-the-art results were produced for narrower grading ranges by training for categorical loss.

To maximize the value of both approaches, models were trained with two objective functions: a classification loss (discussed in 6.3 below) and MeanSquaredError. The hyperparameter $P$ representing the weighting of the classification loss (with $1-P$ as the regression loss) is defined with a logistic function using the essay set metadata:

$$P_{classify} = \frac{L}{1+e^{k(n_c - \overline{n_c})}} + c$$

where $n_c$ is the number of classes in a given essay, $\overline{n_c}$ is the average number of classes across all essays, k is the slope of the logistic function (set to 0.5), and L and c are constants marking the limits of the weighting, set to 0.9 and 0.001 respectively.

### 6.3 Cross-Entropy vs Ordinal Classification Loss

While Categorical Cross Entropy (CCE) is reliable and powerful, the ordinal nature of essay scoring labels offers additional information about each label that can be used when calculating loss as a function related to the closeness to the actual label. A custom objective function based on

QWK[5] to include label smoothing was constructed and the comparison between it and CCE as the objective are shown in in Table 3 for the LSTM model.

| Essay Set | CCE  | Ordinal Loss |
|-----------|------|--------------|
| 1         | 0.67 | 0.72         |
| 2         | 0.62 | 0.58         |
| 3         | 0.66 | 0.69         |
| 4         | 0.79 | 0.81         |
| 5         | 0.79 | 0.77         |
| 6         | 0.74 | 0.74         |
| 7         | 0.70 | 0.74         |
| 8         | 0.29 | 0.42         |
| Avg       | 0.66 | **0.68**     |

Table 2: Best QWK scores for each essay with the LSTM model (no label smoothing)

### 6.4 Embedding-sensitive Optimizer Selection

Selecting to represent all the essay information through sentence-level embeddings requires maximizing the information preserved through embeddings. Small sample sizes and requiring high dropout necessitates an optimizer that provides greater numerical stability. Essays were tested using the Adam and AdaMax optimizers (Kingma et al., 2015).

The AdaMax algorithm implements Adam using the infinity norm as an approximation of $L^2$ norm for high dimensionality (such as embeddings) and results in a learning rate that maximizes the step taken based on the previous step and current gradient:

$$u_t = \max(\beta_2 \cdot u_{t-1}, |g_t|)$$

The difference in performance is illustrated in Figure 3

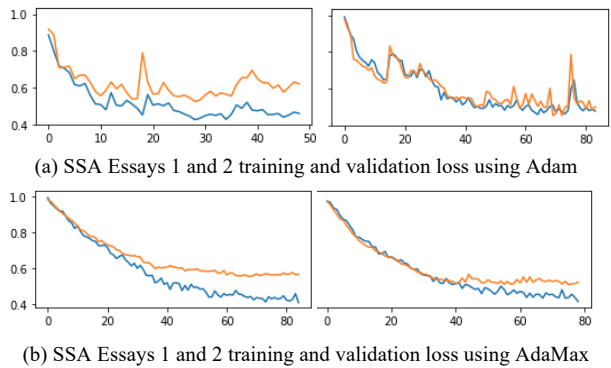

(a) SSA Essays 1 and 2 training and validation loss using Adam

(b) SSA Essays 1 and 2 training and validation loss using AdaMax

Figure 3. Adam vs AdaMax [$\alpha$=0.001, $\beta_1$=0.9, and $\beta_2$=0.999]

### 6.5 Reduced Data Sets

Each of the experiments was conducted multiple times, using various portions of the already-reduced five-fold training and validation sets in order to estimate. While the results are not reported due to space constraints, the overall classification loss to the best performing models stayed above baseline at a 40% reduction in samples.

| Model | 1 | 2 | 3 | 4 | 5 | 6 | 7 | 8 | Avg |
|---|---|---|---|---|---|---|---|---|---|
| LSTM (Alikaniotis) | 0.47 | 0.28 | 0.50 | 0.58 | 0.51 | 0.50 | 0.67 | 0.25 | 0.47 |
| Multi-stage DNN (Jin) | 0.77 | 0.69 | 0.63 | 0.76 | 0.74 | 0.68 | 0.63 | 0.57 | 0.69 |
| EASE (baseline) | 0.76 | 0.61 | 0.62 | 0.74 | 0.78 | 0.78 | 0.73 | 0.62 | 0.71 |
| CNN-RNN (Dasgupta) | 0.80 | 0.63 | 0.71 | 0.71 | 0.80 | **0.83** | 0.82 | 0.70 | 0.75 |
| Human Raters (Shermis 2014) | 0.73 | **0.80** | 0.76 | 0.77 | **0.85** | 0.74 | 0.72 | 0.61 | 0.75 |
| LSTM+CNN Ensemble (Taghipour) | 0.82 | 0.69 | 0.69 | 0.81 | 0.81 | 0.82 | 0.81 | 0.64 | 0.76 |
| LSTM+CNN+Attention (Dong) | 0.82 | 0.68 | 0.67 | 0.81 | 0.80 | 0.81 | 0.80 | 0.71 | 0.76 |
| BERT Multistage Ensemble (Liu) | **0.85** | 0.74 | 0.73 | 0.80 | 0.82 | 0.79 | 0.76 | 0.68 | 0.77 |
| String Kernel / Word Embeds (Cozma) | **0.85** | 0.73 | 0.68 | **0.83** | 0.83 | **0.83** | 0.80 | 0.73 | **0.79** |
| Top Private Combined (Shermis 2014) | 0.82 | 0.74 | 0.75 | 0.82 | 0.83 | 0.81 | **0.84** | 0.73 | **0.79** |
| USE+LSTM | 0.72 | 0.58 | 0.69 | 0.81 | 0.77 | 0.74 | 0.74 | 0.42 | 0.68 |
| USE+ MHA | 0.74 | 0.48 | 0.54 | 0.74 | 0.68 | 0.64 | 0.72 | 0.74 | 0.66 |
| USE+2MHA | 0.75 | 0.74 | 0.68 | 0.83 | 0.78 | 0.79 | 0.77 | 0.62 | 0.74 |
| USE+2MHA+BLSTM | 0.83 | 0.64 | 0.70 | 0.82 | 0.79 | 0.82 | 0.80 | 0.65 | 0.76 |
| PD-S-BERT | - | - | 0.75 | **0.83** | 0.81 | 0.81 | - | - | - |

Table 3: Results reported in QWK for each essay set against current state-of-the-art and baselines.

---

[5] Code used to construct the classifier used the kappa_loss function from tensorflow add-ons: https://github.com/tensorflow/addons/blob/v0.10.0/tensorflow_addons/losses/kappa_loss.py

### 6.6 Transfer Model

The final model, Passage-Dependent Sentence-BERT (PD-S-BERT) does not represent good experimentation, as it incorporates many new tools and methods. Its primary purpose, for the sake of this paper, is to introduce the discussions towards the ideal of having a zero-shot essay grading model that can take a passage context, a learning expectation, a prompt, a rubric, and an essay and from that generate robust insights across multiple tasks related to student literacy. The model is still being refined, but it stands on its own as a step forward in the direction of transfer learning. The technical details can be found in the Appendix.

## 7 Results

The results of the models can be found in Table 3. Listed as producing the top state-of-the-art results is "Top Private Combined" which represents a composite of the highest scores across the key private companies participating in the ASAP challenge (Shermis, 2014). This composite, while it represents older data, is still significant because it represents companies with access to significant amounts of data.

While the more complex attention model with LSTMs neared SOTA results, it had, along with all the models, unusually poor performance on Essay 2. Investigation of losses varies greatly but appears to be connected to the variety of approaches students took when interpreting "library censorship."

However, SOTA performance was achieved on two different essay sets: a literary analysis of author's purpose in a story arc and a narrative on laughter with 60 possible labels. The latter demonstrates the power of using attention on sentence embedding because it was able to capture meaning on the essay set that has half as many observations, is much longer on average, and has dramatically more possible labels than other essays sets. This essay best captures the complexity of creating educator-focused AES systems.

The greatest highlight in the results were in the initial transfer applications of the passage-dependent model. With the passage-dependent model, built for future transfer learning, we see huge gains in performance under this single model, which can perform at SOTA levels given as the input any of the four passage-dependent essays. These results hint at the promise of zero-shot, few-shot, and/or multitask learning models are most exemplified with this model.

### 7.1 Sentence-embeddings as Insights to Reading Comprehension

A criticism of complex language models is that they cannot be used to make inferences about reading or writing. A key benefit of using sentence-embeddings which can capture semantic information is we can now provide, for the first time, insight into a student's reading through analysis of their writing. Using measures of semantic similarity (Cer et al., 2018), each sentence of a student's essay can be linked directly back to the passage: by highlighting sentences with semantic similarity, teachers can see trends across the details in the passage students used to arrive at their conclusions. Figure 4 contains a screenshot of the semantic similarity calculation used to highlight the most semantically related sentences to the selected sentence from the student response, where the saturation of the highlight is proportional to the magnitude of the similarity.

*Figure 3: Screenshot of a functioning user interface built to demonstrate the use of sentence embeddings with semantic information to highlight details from a source passage that are related to a sentence found within an individual student response.*

## 8 Conclusion

These modest results provide optimism for the use of semantic embeddings, transformer attention, and custom objective functions and hyperparameters as foundations for future research towards more meaningful educator-focused AES systems.

The work of this paper only scratches the surface of the tremendous work needed to move automated essay scoring from the realms of businesses and academic exercises into the hands of teachers and students. Much work is left to be done, but these results promise exciting possibilities for future work that could enhance the education experiences for public school students everywhere. Having free, accurate pedagogically aligned automatic scoring options will lead to increases of more authentic assessments for improving student learning.

## Acknowledgements

Thanks to Dr. Daniel Cer for guidance in interpretation of sentence embeddings and recommendations for future avenues of research.